\documentclass[10pt,twocolumn,letterpaper]{article}

\usepackage{iccv}
\usepackage{times}
\usepackage{epsfig}
\usepackage{graphicx}
\usepackage{amsmath}
\usepackage{amssymb}

\usepackage{algorithm}
\usepackage{algorithmic}

\usepackage[export]{adjustbox}
\usepackage{makecell}
\usepackage{siunitx}
\usepackage{multirow}

\usepackage{subcaption}
\usepackage{graphicx}
\usepackage{wrapfig}
\usepackage[dvipsnames]{xcolor}

\iffalse % don't show comments
    \newcommand\todo[1]{}
    \newcommand{\jiaman}[1]{}
    \newcommand{\ruben}[1]{}
    \newcommand{\duygu}[1]{}
    \newcommand{\jimei}[1]{}
    \newcommand{\zhengfei}[1]{}
\else % do not show comments
    \newcommand{\todo}[1]{{\textcolor{red}{[[TODO: {#1}]]}}}

    \newcommand{\zhengfei}[1]{\textcolor{Aquamarine}{[Zhengfei: {#1}]}}
    \newcommand{\jiaman}[1]{\textcolor{red}{[Jiaman: {#1}]}}
    \newcommand{\ruben}[1]{\textcolor{magenta}{[Ruben: {#1}]}}
    \newcommand{\duygu}[1]{\textcolor{green}{[Duygu: {#1}]}}
    \newcommand{\jimei}[1]{\textcolor{blue}{[Jimei: {#1}]}}
\fi

\iftrue % use space-saving macro

\else % do not use space-saving macro

\fi

\makeatletter

\def\etal{\emph{et al.}}
\makeatother

\usepackage[pagebackref=true,breaklinks=true,letterpaper=true,colorlinks,bookmarks=false]{hyperref}

\iccvfinalcopy % *** Uncomment this line for the final submission

\def\iccvPaperID{2800} % *** Enter the ICCV Paper ID here

\ificcvfinal\fi

\begin{document}

\linespread{0.9}

%%%%%%%%% TITLE
% \title{Learning Variational Priors for Human Motion Restoration}
%\title{Hierarchical Variational Autoencoders for Human Motion Restoration}
\title{Task-Generic Hierarchical Human Motion Prior using VAEs}

\author{Jiaman Li$^{1,2}$,\enspace Ruben Villegas$^3$,\enspace Duygu Ceylan$^3$,\enspace Jimei Yang$^3$, \\
\enspace Zhengfei Kuang$^{1,2}$, \enspace Hao Li$^{4,5}$, \enspace Yajie Zhao$^2$ \\ \\
$^1$University of Sounthern California\enspace$^2$USC Institute for Creative Technologies \\  
\enspace
$^3$Adobe Research\enspace$^4$Pinscreen\enspace$^5$UC Berkeley \\
% {\tt\small \{jiamanli, zkuang\}@usc.edu\enspace hao@hao-li.com}
}

% \maketitle

\twocolumn[{%
\renewcommand\twocolumn[1][]{#1}%
% \vspace{-3em}
\maketitle
\vspace{-8mm}
\begin{center}
    \centering
    \includegraphics[width=\textwidth]{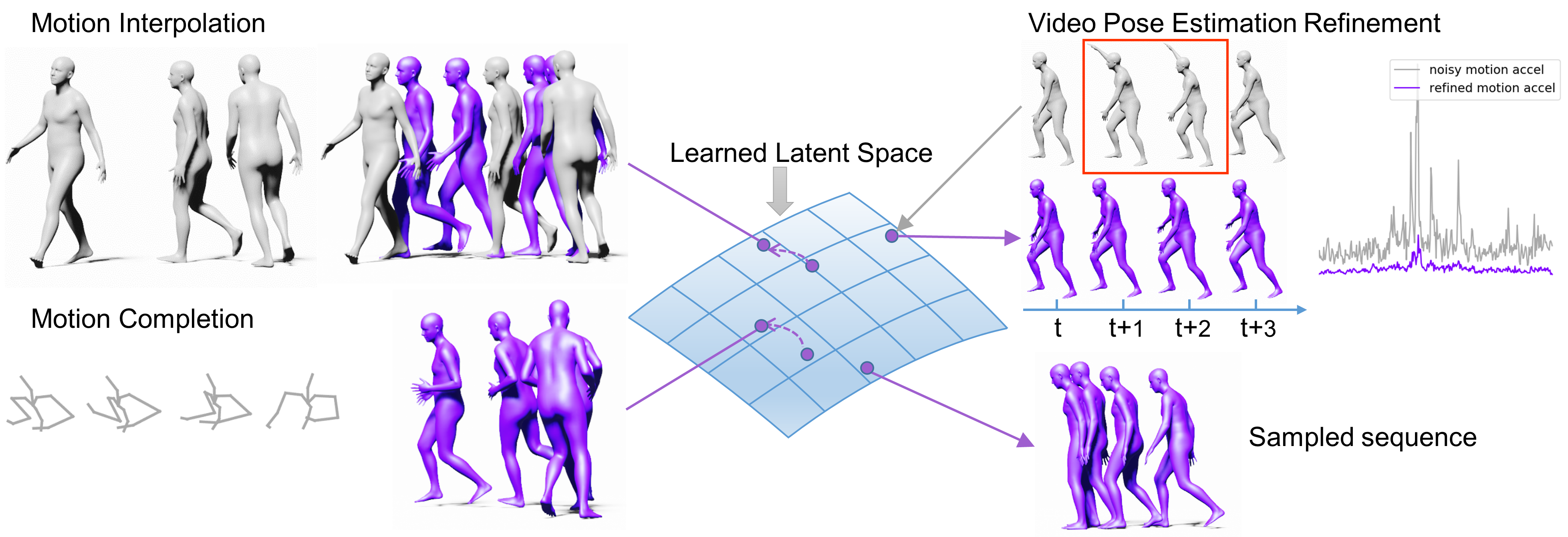}
    \vspace{-5mm}
    \captionof{figure}{ Our general purpose motion prior consists of a latent space of human motions and is learned using a hierarchical motion variational autoencoder (HM-VAE). Our approach is task-generic and can be directly adopted to a wide range of applications. \textit{Left:} Motion interpolation and completion can be accomplished by traversing the latent space. \textit{Right:} Noisy pose estimation can be refined by projecting noisy inputs into our latent space and decoding back. And a latent vector in the learned latent space is corresponding to a valid motion sequence.}
    \label{figure1}
    %\vspace{-5pt}
\end{center}%
}]

% \twocolumn[{%
% % \renewcommand\twocolumn[1][]{#1}%
% \maketitle
% \vspace{-8mm}
% \begin{center}
%     \centering
%     % \begin{figure}
%     \includegraphics[width=\textwidth]{latex/figures/Figure_1.png}
%     \label{figure1}
%     % \end{figure}
% \end{center}%
% \vspace{-3mm}
%     % \captionof{Figure 1: }{The learned motion prior forms a manifold. \textit{Left:} Motion interpolation/completion can be accomplished by traversing the latent space. \textit{Right:} Noisy motions can be fixed by projecting to latent space and decode.}
%     \captionof{Figure 1: }{Human motion latent space learned by our hierarchical VAE. \textit{Left:} Motion interpolation and completion can be accomplished by traversing the latent space. \textit{Right:} Motion denoising is achieved by projecting noisy inputs into our latent space and decoding back.}
%     \vspace{4mm}
%     % \caption{We learn motion prior from high-quality motion capture data. The learned motion latent space is constrained to a manifold. Each point in the manifold is corresponding to a motion sequence with a fixed timestep window. Motion completion and interpolation task can be accomplished by traversing the latent space given target joint rotations or key frame rotations. Also, by projecting noisy motions to our latent space and decode from latent vector, we are able to get smooth motions in better quality.}
% % \end{center}%
% }]

% \maketitle
% Remove page # from the first page of camera-ready.
\ificcvfinal\thispagestyle{empty}\fi

\setcounter{figure}{1}

%%%%%%%%% ABSTRACT
% \vspace{8mm}
\begin{abstract}
% \vspace{-4mm}
A deep generative model that describes human motions can benefit a wide range of fundamental computer vision and graphics tasks, such as providing robustness to video-based human pose estimation, predicting complete body movements for motion capture systems during occlusions, and assisting key frame animation with plausible movements. In this paper, we present a method for learning complex human motions independent of specific tasks using a combined global and local latent space to facilitate coarse and fine-grained modeling. Specifically, we propose a hierarchical motion variational autoencoder (HM-VAE) that consists of a 2-level hierarchical latent space. While the global latent space captures the overall global body motion, the local latent space enables to capture the refined poses of the different body parts.
We demonstrate the effectiveness of our hierarchical motion variational autoencoder in a variety of tasks including video-based human pose estimation, motion completion from partial observations, and motion synthesis from sparse key-frames. Even though, our model has not been trained for any of these tasks specifically, it provides superior performance than task-specific alternatives. Our general-purpose human motion prior model can fix corrupted human body animations and generate complete movements from incomplete observations. 
% By projecting initial predictions to our learned latent space, we show that we improve the video-based human estimation methods in terms of temporal coherency while preserving the motion quality.
% Our prior also effectively performs motion completion and motion interpolation.
\end{abstract}

%%%%%%%%% BODY TEXT
\vspace{-4mm}
\section{Introduction}

% Story 1: From the perspective of learning priors itself: We propose a variational motion priors learning from mocap data and found that the learned priors are useful in different motion tasks even we did not design for each task. 
% P1: Motion is important. 
% P2: Prior work.
% P3: Our work.
% Summarize our contribution.  

% (Deprecated?) Story 2: From the perspective of obtaining high-quality motion, key-frame based animation by artists and mocap are both expensive, limited to the motion type and scale, a promising direction is to extract high-quality motion from videos and automatically generate in-betweening frames instead of manual work by animators. However, usually motions from videos are noisy, some have jitters and sudden pose changes, we propose a general motion prior learning approach, the learned priors can be used in different scenarios when the motion has low-quality. Slerp-based interpolation is unable to address longer intervals. We show two specific scenarios, one is denoising video pose estimation results, the other is filling in missing frames given key frames. 

% Why do we do this motion prior instead of treating each task independently? 
% Why do we use VAE?
% Why do we use skeleton-aware model? 

% (Background and Application of high-quality 3d motion)
% List some scenarios that we need motion restoration? 
% Video pose estimation, not smooth, quality is not good. 
% Motion interpolation, slerp is not good for large intervals. 
% IK: not work well? 
% Humans has the ability to imagine complete motion with only observe partial motions? 

The modeling of human motions is a core component for many vision tasks, including pose estimation, action recognition, motion synthesis, and motion prediction. Several recent work have demonstrated new capabilities for generating complex body movements and capturing motion from unconstrained videos~\cite{kocabas2020vibe,kanazawa2019learning,zanfir2020weakly,arnab2019exploiting,dabral2018learning,rayat2018exploiting,kanazawa2018end,pavllo20193d}. While robustness and accuracy is constantly evolving for these methods, highly challenging scenes, occlusions, and body poses can still result in corrupted animations and noise. 

Conventional techniques for reducing artifacts, include temporal filtering~\cite{kalmanfilter}, inverse kinematics~\cite{welman1993inverse,rose2001artist,grochow2004style} and statistical human motion priors~\cite{wang2007gaussian,ikemoto2009generalizing,lehrmann2014efficient,akhter2015pose}. While effective in reducing unwanted jitters and implausible poses, these methods do not generalize well to complex motions and the results are often inaccurate w.r.t. the ground truth.

To address this challenge, deep learning-based motion priors were proposed which are particularly effective in representing complex motion variations ~\cite{holden2015learning,holden2016deep,kocabas2020vibe,luo20203d}. These priors are generally designed for predetermined tasks, such as 3D pose estimation from a video, and a common problem is to be able to cover all possible input cases during training, such as occlusions, motion blur, etc. Ideally, we could build a prior model that describes the space of plausible human body movements, independently of the application and simply plug this model into any system. Training such model would simply consist of collecting high-quality motion capture data (task-generic), instead of fitting for example 3D models to an image (task-specific).

We introduce a generalized motion prior, that learns complex human body motions from high-fidelity motion capture data~\cite{mahmood2019amass}. Similar to the work of~\cite{yang2020deep} who developed a deep optimized prior for 3D modeling, our prior for motion is a general purpose one.
We present a deep generative model based on a joint global and local latent space representation that can accurately capture the poses of different body parts while also modeling the global correlations across the body joints. Specifically, we adopt a two-level hierarchical motion variational autoencoder (HM-VAE) which maps the human motion to global and local latent spaces simultaneously. Our HM-VAE model adopts the recently proposed skeleton-aware architecture~\cite{aberman2020skeleton} and defines the global and local latent spaces via direct pooling and unpooling operations on the skeleton structure. 
% Our method uses a pre-defined time window of frames, which can be specified by a user, to infer a corrected frame.
While our HM-VAE successfully models the local human motion, we introduce an additional trajectory prediction component to model global motions.
% \ruben{The main point we should emphasize is the motion priors. The trajectory prediction component is only necessary because we want to simulate motion globally. We need to be careful about this to not confuse reviewers because they will ask us to use the trajectory prediction in the video pose estimation task and I think we aren't doing this.}. \jiaman{Trajectory prediction model is still a new component, I think we should mention it just for a couple of sentences, not to emphasize it.}
Taking local joint positions as input, our trajectory model estimates the root joint velocity at each timestep, enabling us to recover human motions in world space.
% \jiaman{(Maybe we should remove this sentence, since this strategy is only used in pose estimation. For motion completion and interpolation, we directly do optimization for long sequence.) During training, our model observes a fixed window of motion.
% Therefore, we introduce a center-frame sliding window strategy to apply our model to arbitrarily long sequences.}s
% \ruben{The sliding window approach is something commonly used. I am not sure if it's a good idea to highlight it as contribution}.
% \jiaman{I think we should mention, that's how we can deal with arbitrary long sequences which MEVA cannot do. MEVA shows obvious discontunity between consecutive windows.}

We show the generality and effectiveness of our human motion prior on various applications.
% First, we show that our model can take noisy human motion predicted from video~\cite{kocabas2020vibe,kanazawa2019learning} as input, map it to the learned latent spaces, and decode high quality motion that both preserves the original motion content and significantly improves the temporal coherence of motion.
First, we show that our task-generic model can refine human motions predicted from video~\cite{kocabas2020vibe,kanazawa2019learning} by mapping noisy predictions into our motion prior latent space.
We also demonstrate that our model can perform motion completion given partial observations (e.g., the upper body motion only) or motion synthesis given sparse keyframes.
In both of these tasks, we optimize for both the global and local latent variables to match the partial observations and restore complete plausible motion sequences.
While our model is not trained for any of these tasks specifically, it outperforms task-specific alternatives both qualitatively and quantitatively.

Our contributions are as follows. First, we present an effective task-generic motion prior model, that can improve the performance of a wide range of applications. Second, we propose a two-level hierarchical motion variational autoencoder (HM-VAE) that consists of a skeleton-aware architecture, allowing it to accurately capture the local motion of body parts and the global correlation between them. Finally, we introduce a trajectory prediction module to model the global trajectory conditioned on the local body motion.

\section{Related Work}
% We review works most related to ours including deep prior learning in other domains, generative motion modeling and motion estimation from videos. \ruben{This sentence is not necessary}  

\paragraph{Deep Learning Based Priors.}
The ability of deep neural networks to model data priors has sparked research in a variety of domains.
Deep Image Prior (DIP)~\cite{ulyanov2018deep} shows that a generator network without any learning is an effective prior for image restoration. Given randomly initialized weights, the neural network is able to perform image denoising or super resolution via optimization defined by a task-dependent energy term and a regularizer.
% \jimei{de-emphasize the un-trained part}
A similar idea is proposed and validated in the video domain~\cite{lei2020blind}, by training a network to mimic specific image operators in a single test video, the learned video prior is able to eliminate temporal inconsistencies in various video processing tasks. Besides discovering priors in the 2D domain, the ability to capture 3D priors is also demonstrated in recent works~\cite{hanocka2020point2mesh,yang2020deep}. Point2Mesh~\cite{hanocka2020point2mesh} randomly samples a fixed vector and optimizes the network parameters to reconstruct a mesh with geometric details and showcases the effectiveness of self-prior. Deep Optimized Priors~\cite{yang2020deep} propose to learn a pre-trained prior first which then serves as initialization for optimizing both the latent vector and the decoder parameters given a task-specific objective and regularization loss. In this work, we investigate data-driven priors in the human motion domain and validate the effectiveness of our method by applying it to various human motion tasks without explicitly training for any specific tasks. 

\paragraph{Generative Motion Modeling.}
With the recent success of learning based methods, several works have focused on generative models for motion synthesis. Martinez \etal~\cite{julieta2017motion} propose a recurrent neural network model for generating future human motion by predicting future joint velocities and adding them to previous joint positions. MT-VAE~\cite{Yan_2018_ECCV} propose a probabilistic recurrent neural network method for generating multiple future human motions. Aksan \etal~\cite{Aksan_2019_ICCV} propose to predict future human motion by exploiting the kinematic structure in human bodies. Following the auto-regressive generative model formulation, Motion Transformers~\cite{li2020learning} are introduced to model the future pose distribution along with a discrete pose representation, leveraging the advantage of the Transformer~\cite{vaswani2017attention} architecture.
MotionVAE~\cite{ling2020character} models the future pose distribution given previous pose using a variational autoencoder (VAE)~\cite{kingma2013auto} approach. Normalizing flows is another category of generative models recently applied to human motion modeling. MoGlow~\cite{henter2019moglow} uses normalizing flows for motion modeling and achieve realistic motion generation taking root trajectory as the conditioning signal. Recent work also address the problem of motion in-betweening from a generative modeling perspective. Long-term motion in-betweening~\cite{zhou2020generative} uses a generative adversarial neural network (GAN)~\cite{goodfellow2014generative} approach to generate human motion given sparse key-frames. Robust Motion in-betweening~\cite{harvey2020robust} employs an LSTM to generate a motion sequence given initial frames and end frames while also allowing for motion variations. Here, we focus on extending VAEs to model long-term human motion. Our key difference is to embed multiple frames of motions into a hierarchical global and local latent spaces. Methods like MotionVAE model motion in a per-frame basis while our method maps a full motion sequence into a compact latent space.
% \jiaman{Instead of modeling latent space per timestep condition on previous pose like MotionVAE, we learn a latent space for the whole motion sequence, enabling a compact representation for motions which facilitate directly motion refinement and optimization.} \jimei{Clarify the difference with motionVAE}

% ~\cite{wang2020synthesizing} synthesis long-term human interaction in scenes. 
% ~\cite{xu2020hierarchical} hireachical style-based motion synthesis 
% pose prior~\cite{lehrmann2013non} Non-parametric Bayesian Networks 

\paragraph{Motion Estimation From Videos.}
%Extracting human motion from videos is a long-standing problem and extensively explored by both non-learning~\cite{anguelov2005scape,urtasun2006temporal,wei2009modeling,wei2010videomocap} and learning-based methods~\cite{guler2019holopose,tekin2016structured,mehta2017vnect,pavllo20193d,shi2020motionet}. %Various approaches have been investigated for 3D pose estimation since deep learning showed great potential in image related tasks. 
%In addition to pose estimation, several works~\cite{anguelov2005scape,loper2015smpl,pavlakos2019expressive} focus on joint body pose and shape regression. 
A multitude of optimization-based~\cite{loper2015smpl,bogo2016keep,huang2017towards}, learning-based~\cite{tan2017indirect,kanazawa2018end}, and hybrid methods~\cite{kolotouros2019learning,pavlakos2019expressive} have been proposed to tackle the problem of single-image 3D human pose estimation. 
%Follow-up works have also investigated the case when the human body is only partially visible~\cite{rockwell2020full}. 
Its rapid progress has stimulated research interest on the long-standing problem of extracting 3D human motion from videos~\cite{anguelov2005scape,urtasun2006temporal,wei2009modeling,wei2010videomocap,guler2019holopose,tekin2016structured,mehta2017vnect,pavllo20193d,shi2020motionet}.
VIBE~\cite{li2020learning} uses an LSTM to capture temporal information and introduce a discriminator training strategy to ensure the predicted poses lie in a valid manifold. MEVA~\cite{luo20203d} presents a coarse-to-fine strategy where a valid motion sequence is first extracted conditioned on a latent vector and then refined utilizing person-specific details. TCMR~\cite{choi2020beyond} focuses on avoiding the temporal jitters that exist in the VIBE results and proposes a strategy to explicitly leverage past and future frames to achieve smoother results. Texture-based tracking~\cite{xiang2019monocular} is shown to improve the motion stability during optimization. Foot contacts and physically-based models~\cite{rempe2020contact} are also used for estimating realistic human motions from videos. In this work, we are not aiming to design a specific 3D video pose estimation method. Instead, we show that our human motion prior is capable of eliminating jitters and noises that exist in the results of current state-of-the-art methods. We demonstrate that our method can be applied to any pose estimation methods and in our experiments we outperform previous work both quantitatively and qualitatively.

\section{Hierarchical Motion VAEs}
% In this section, we first review our deployed backbone Skeleton-Aware Model, then introduce our approach for motion priors learning. 
The core of our method is a hierarchical motion variational autoencoder (HM-VAE) that models human motion by jointly learning a local and global latent space. Specifically, given a motion sequence $ \mathbf{x} \in R^{T \times J \times D}$, represented as the $D$ dimensional joint rotations in a fixed time window of size $T$\footnote{In our experiments, we use the SMPL~\cite{loper2015smpl} skeleton hence the number of joints $J$ is 24 and we use the continuous 6D rotation representation~\cite{zhou2019continuity}, hence $D=6$.}, we first learn an embedding of $\mathbf{x}$ into local and global latent spaces represented by latent codes $z_{l}$ and $z_{g}$ respectively. Assuming the latent space in the local and global levels are independent~\cite{li2020progressive}, we then model the probability distribution of a motion sequence as:
% $$
% p(x, z) = p(x|z_{l}, z_{g})p(z_{l})p(z_{g}))
% $$
\begin{equation}
    p(\mathbf{x}, z) = p(x|z_{l}, z_{g})p(z_{l})p(z_{g}).
\end{equation}

Our variational autoencoder adopts the recently proposed skeleton-aware architecture~\cite{aberman2020skeleton} to facilitate learning over the humanoid skeleton structure directly. Before we discuss the details of our model, we first provide a brief overview of the skeleton-aware architecture. We refer the reader to the original paper for more details.

%\ruben{Should we consider removing background outside of the method section?}
%\duygu{We could move background and overview to a separate section that comes before the method. We can keep the description of skeleton aware network shorter.}
\subsection{Background} 
The skeleton-aware architecture consists of three critical components that we adopt in our model design: skeleton convolution, skeleton pooling, and skeleton unpooling.

% We will describe each operation as follows.
\paragraph{Skeleton Convolution.}
% \zhengfei{this part is a bit hard to understand. It might be better to denote the skeleton as $S$ in the beginning, then say $\forall i \in S$, we calculate $y_i$ by blabla. Also, there's no explanation between $T, k$ and $T'$} 
Given a motion sequence $\mathbf{x}, \mathbf{x} \in R^{T \times J \times D}$, we denote $\mathbf{y}, \mathbf{y}\in R^{T' \times J \times D'}$ as the updated features after a skeleton convolution operation.
For each bone $i$ in the skeleton, the updated feature is calculated as $\mathbf{y}_i = \frac{1}{|N^d_i|} \sum_{j\in N^d_i}{\mathbf{x}_j*W_j^i+b_j^i}$,
% \zhengfei{are you sure the index for bias is $i$ and $j$? You can merge all bias together anyway.} 
where the symbol $*$ denotes a one dimensional temporal convolution operation with the temporal filter $W_j^i \in R^{k \times D \times D'}$ and bias $b_j^i \in R^{D'}$. $D'$ represents the number of temporal filters, $k$ represents the temporal kernel size, and $N^d_i$ represents the neighboring bones of bone $i$ within distance $d$. The distance between two bones $(j_1, j_2)$ is defined as the number of bones needed to cross to reach $j_2$ starting from $j_1$ along the kinematic chain. The skeleton convolution operation preserves the number of edges $J$ while downsampling the temporal dimension to $T'$.

\paragraph{Skeleton Pooling.}
% Similar to pooling operation in model for processing images, skeleton pooling is introduced to merge adjacent joint information and extract high-level motion features.
Skeleton pooling merges the features of connected bones and extracts higher-level motion features by reducing the spatial resolution of the input. The pooling operation is applied to pairs of bones which are connected by a joint with degree of 2. For example, the thigh and calf which are connected by the knee. We recursively search such bone pairs starting from the root node (the hip), and merge their corresponding features using average pooling operation. As we perform pooling, the number of joints is reduced in subsequent layers of the network. Given disjoint sets of pooling bones denoted as $\{P(1), P(2), ..., P(m)\}$, the pooling operation is defined as 
{$$ F'_i = \text{Pool}(F_j|j \in P(i)). $$}

\vspace{-8mm}
\paragraph{Skeleton Unpooling.} The unpooling operation mirrors skeleton pooling. Specifically, given the activation features $F$ defined on a bone $b$ obtained by merging the bones $(i,j)$, unpooling simply replaces $b$ with the bones $i$ and $j$  where the new bone features are defined as $F_i = F, F_j = F$. The number of bones is increased after the unpooling layer.  % \zhengfei{the last sentence can be removed.}
% \subsection{Problem Formulation}

\begin{figure*}[h]
\begin{center}
\vspace{-5mm}
\includegraphics[width=0.9\textwidth]{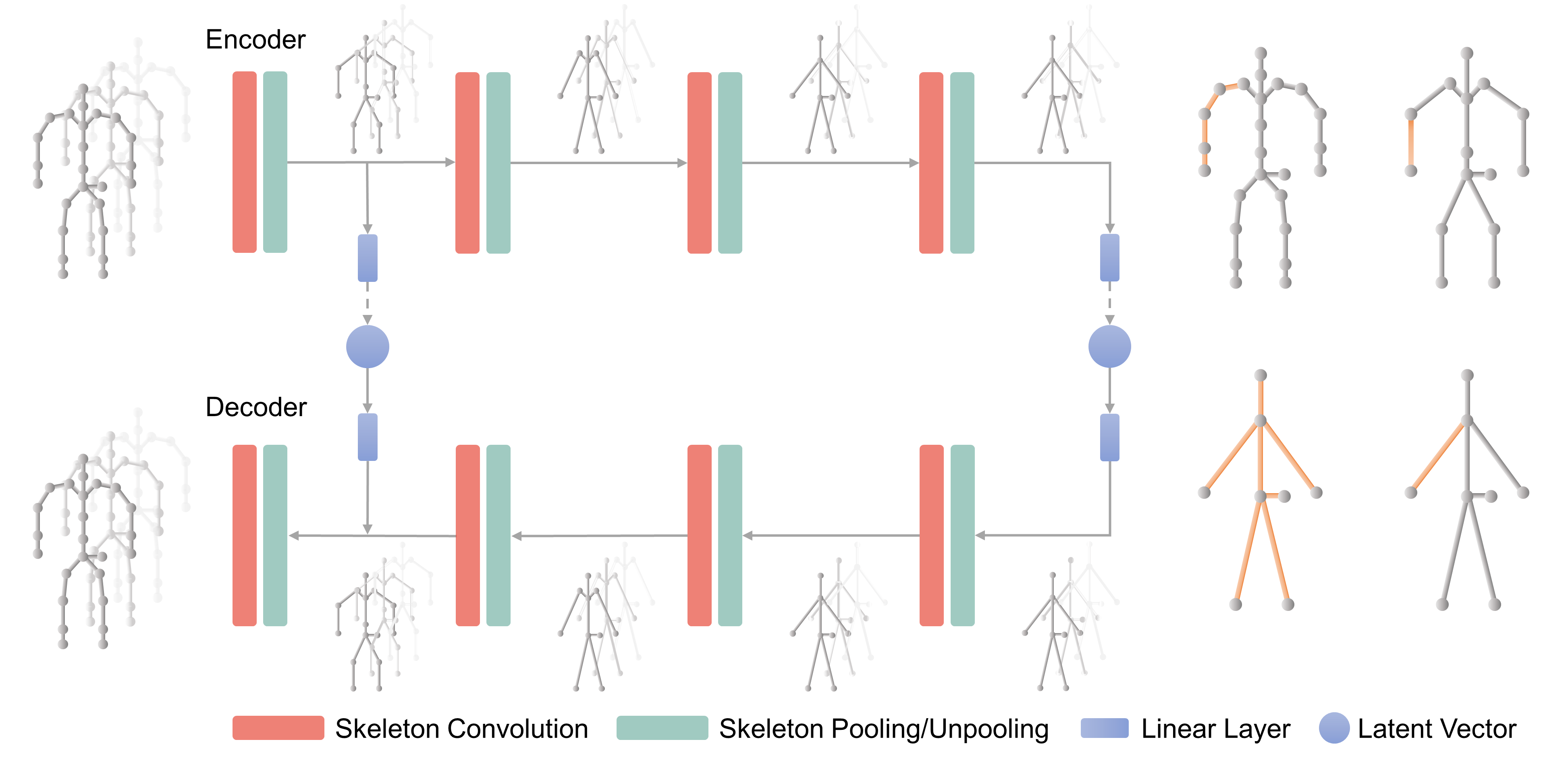}
\end{center}
\vspace{-6mm}
\caption{Model Overview. \textit{Left:}  Model architecture. (We omit activation layers and temporal upsampling layers here for simplicity.) \textit{Right:} Illustration of receptive field in shallow ($B_1$) and deep layers ($B_4$).}
\label{figure_model}
\vspace{-1mm}
\end{figure*}

\subsection{Motion Prior Learning for Local Motion}
Given a motion sequence $\mathbf{x}, \mathbf{x} \in R^{T \times J \times D}$, our HM-VAE consists of an encoder and a decoder as shown in Figure~\ref{figure_model}. The encoder learns the posterior distribution of the local, $z_{l}$, and global, $z_{g}$ latent spaces given data $\mathbf{x}$: 
% $$
% q(z_{l}, z_{g}|x) = q(z_{l}|f_{l}(x))q(z_{g}|f_{g}(x))
% $$
\begin{equation}
    q(z_{l}, z_{g}|x) = q(z_{l}|f_{l}(x))q(z_{g}|f_{g}(x)),
\end{equation}
where $f_{l}(x)$, $f_{g}(x)$ represent the motion features extracted from different layers in the encoder. Our VAE is then trained by maximizing the modified Evidence Lower BOund (ELBO)~\cite{kingma2013auto}:

% Given $z_{l}, z_{g}$, the decoder reconstructs the motion sequence $\mathbf{x}$.
% The VAE is trained by minimizing the ELBO~\cite{kingma2013auto}: 
% $$
% \log p(x) \geq \mathbb{E}_{q(z|x)}[\log p(x|z)] - \beta KL(q(z|x)||p(z))
% $$
% \begin{align}
%     \log p(x) \geq \ & \mathbb{E}_{q(z_l,z_g|x)}[\log p(x|z_l,z_g)] - \nonumber \\
%     & \beta KL(q(z_l,z_g|x)||p(z)).
% \end{align}
% \jiaman{

\begin{align}
    \log p(x) \geq \ & \mathbb{E}_{q(z_l,z_g|x)}[\log p(x|z_l,z_g)] - \nonumber \\
    & \beta KL(q(z_l|f_{l}(x))||p(z_l)) - \nonumber \\
    & \beta KL(q(z_g|f_{g}(x))||p(z_g)),
\end{align}
% \begin{align}
%     \log p(x) \geq \ & \mathbb{E}_{q(z|x)}[\log p(x|z)] - \nonumber \\
%     & \beta KL(q(z|x)||p(z))
% \end{align}
% }
where $q(z_l,z_g|x)$ is an encoder network that maps the input $x$ into the local and global latent spaces, $p(x|z_l,z_g)$ is a decoder network that maps latent variables back into the input $x$, and $p(z_l)$ and $p(z_g)$ are assumed to be standard normal distributions $\mathcal{N}(0, \mathcal{I})$.

% \duygu{how is p(z) defined?}
% \jiaman{$p(z_l)$, $p(z_g)$ is prior distribution of latent variables and is set to a normal distribution $\mathcal{N}(0, \mathcal{I})$.}
\vspace{-2mm}
\paragraph{Encoder.}
The encoder consists of four building blocks $B_1, B_2, B_3, B_4$ where each building block is a combination of a skeleton convolution, skeleton pooling, and a LeakyReLU activation layer. As shown in Figure~\ref{figure_model}, we introduce a linear layer $W \in R^{T'd \times 2d_h}$ after $B_1$ and $B_4$, mapping motion features of each corresponding block to a latent space. While the shallow layer features $F_l$ after $B_1$ represent the local latent space, the deep layer features $F_g$ after $B_4$ correspond to the global latent space. We enforce a normal distribution on each latent space:
% $$z_l\sim \mathcal{N}(\mu_l(F_l), \sigma_l(F_l)), z_g\sim \mathcal{N}(\mu_g(F_g), \sigma_g(F_g)) $$
\begin{equation}
    z_l\sim \mathcal{N}(\mu_l(F_l), \sigma_l(F_l)), z_g\sim \mathcal{N}(\mu_g(F_g), \sigma_g(F_g)).
\end{equation}

%We employ the Skeleton-Aware Model architecture in encoder. By taking $X, X\in R^{T \times J \times D}$ as input to Skeleton-Aware Encoder which consists of four skeleton convolution and skeleton pooling layers along with PReLU activation layers following pooling operations, we get the updated motion features $F, F\in  R^{J' \times T' \times d}$. We denote a computation block as a combination of one skeleton convolution layer, one skeleton pooling layer and one activation layer, then we have $B_1, B_2, B_3, B_4$ which represents 4 blocks respectively. 

\vspace{-6mm}
\paragraph{Decoder.}
The decoder has a symmetric architecture to the encoder. Each building block in decoder consists of temporal upsampling, skeleton unpooling, skeleton convolution and LeakyReLU activation layers. Given the latent codes $z_l$ and $z_g$, the decoder first maps them to features through linear layers. Temporal upsampling and skeleton unpooling operations are used to increase the number of timesteps and bones gradually. The features obtained from the global latent code after a series of temporal upsampling, skeleton unpooling and convolution are concatenated with the features obtained from the local latent code. A final block of unpooling and convolution operations are used to reconstruct the original motion sequence $\mathbf{x}$. We further add a forward kinematics layer proposed in ~\cite{villegas2018neural} to convert $\mathbf{x}$ into joint positions $\mathbf{P}$ to define an additional joint position reconstruction loss. Also, we convert the 6D rotation representation to the rotation matrix $\mathbf{R}$ and use an additional reconstruction loss defined on the rotation matrices. Overall, the reconstruction loss used to train the decoder is defined as:
% $$ L_{rec} = L_{6d} + L_{rot} + \lambda L_{pose}$$
% $$ L_{6d} = \sum_{t=1}^{T} ||X'_{t}-X_{t}||^2$$
% $$ L_{rot} = \sum_{t=1}^{T} ||R'_{t}-R_{t}||^2$$
% $$ L_{pose} = \sum_{t=1}^{T} ||P'_{t}-P_{t}||^2$$
\begin{align}
    L_{rec} &= L_{6d} + L_{rot} + \lambda L_{joints}, \\
    L_{6d} &= \sum_{t=1}^{T} ||\mathbf{x}'_{t}-\mathbf{x}_{t}||^2, \\
    L_{rot} &= \sum_{t=1}^{T} ||\mathbf{R}'_{t}-\mathbf{R}_{t}||^2, \\
    L_{joints} &= \sum_{t=1}^{T} ||\mathbf{P}'_{t}-\mathbf{P}_{t}||^2.
\end{align}
we experimentally set $\lambda$ to 10 in our training process.  % \ruben{We usually call rotations "pose" so we should probably change $L_{pose}$ so something like $L_{joints}$}

% Face compositionalVAE~\cite{bagautdinov2018modeling}

% \begin{figure*}[h]
% \begin{center}
% \vspace{-5mm}
% % \includegraphics[width=100pt,height=25pt]{figures/model_pipeline.pdf}
% \includegraphics[width=0.5\textwidth]{latex/figures/Figure_2_Model.png}
% \end{center}
% \vspace{-4mm}
% \caption{Model (Placeholder)}
% \label{model_pipeline}
% \vspace{-1mm}
% \end{figure*}

% \begin{figure}[t!]
% \vspace{-2mm}
% \begin{center}   \includegraphics[width=\linewidth]{latex/figures/Figure_2_Model.png}
% \end{center}
% \vspace{-5mm}
% \caption{Model (placeholder)}
% \label{figure_model}
% \vspace{-3mm}
% \end{figure}

\subsection{Trajectory Prediction} 
% \ruben{This is not part of the prior. This is in fact useful, but I am not sure it's novel or needs its own subsection. We need to present it as a tool to get global translation from our motion prior} \duygu{I think it's still useful to present it here, without this component we wouldn't be able to model global motion.} \ruben{If we add it as part of the method, will reviewers ask us to also present results from trajectory prediction in the pose estimation experiments?}
Given a motion sequence, we use the presented HM-VAE to model the local motion, i.e., the local joint rotations. In addition, we utilize a similar skeleton-aware architecture without reducing temporal dimension to model the global root joint trajectory. Specifically, given a sequence of joint positions denoted as $\mathbf{P} \in R^{T \times J \times 3}$, we apply four skeleton convolution layers with skeleton pooling layers to obtain motion features $F \in R^{T \times J' \times d}$. We use a linear layer that takes $F$ as input and estimates the root velocity $V \in R^{T \times 3}$. By accumulating the root velocity in subsequent frames, we compute the global root trajectory $G \in R^{T \times 3}$. The root trajectory at any particular time $t$ is defined as $G_t = \sum_{i=0}^{t} V_i$. We train the trajectory estimation module with a loss function that consists of both velocity and trajectory terms: 
% $$ L_{traj} = \sum_{t=1}^{T}||V_t - V_t'||^2 + ||G_t - G_t'||^2 $$
\begin{equation}
    L_{traj} = \sum_{t=1}^{T}||V_t' - V_t||^2 + ||G_t' - G_t||^2 
\end{equation}

\begin{figure*}[t!]
\vspace{-2mm}
\begin{center}   \includegraphics[width=0.9\linewidth]{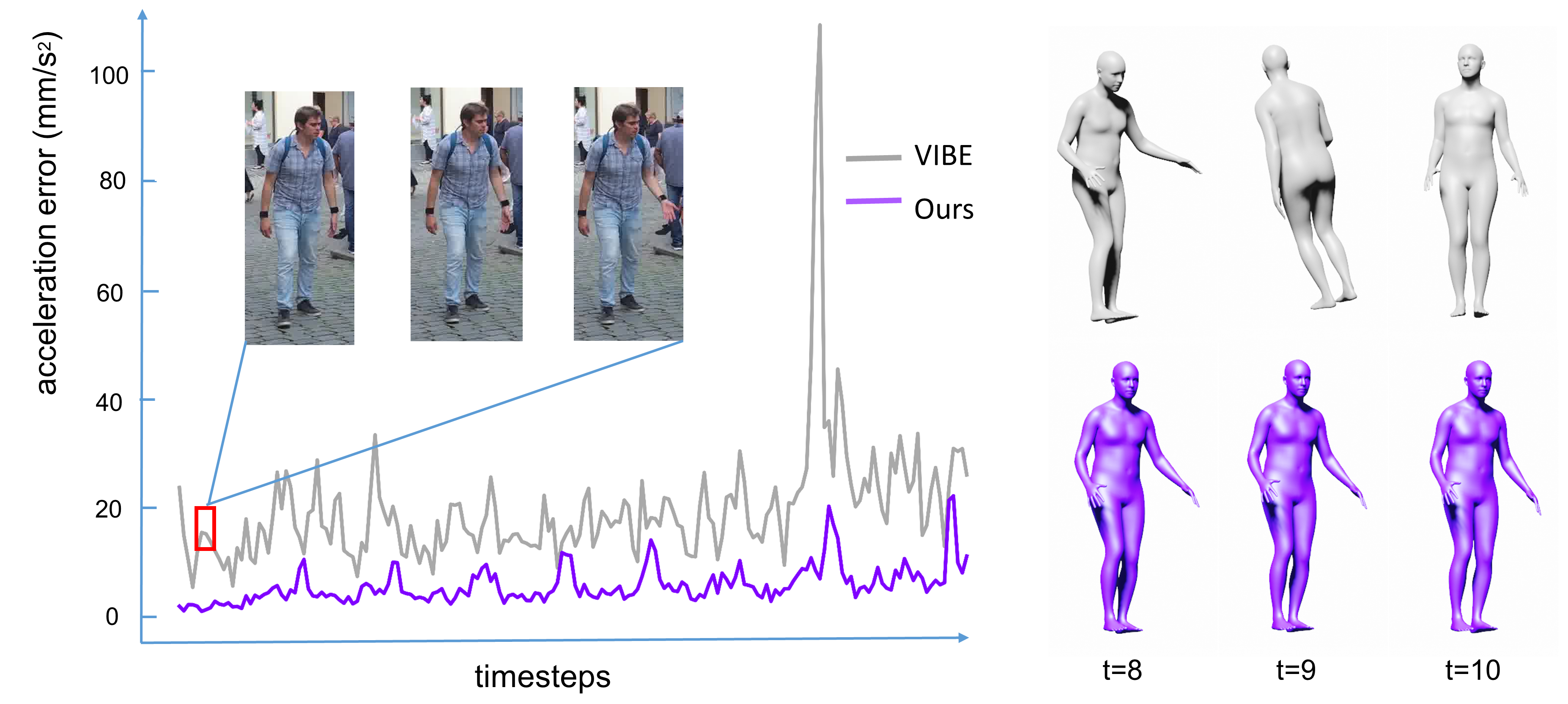}
\end{center}
\vspace{-5mm}
\caption{Acceleration error curves for VIBE~\cite{kocabas2020vibe} results and our refined results. The right figure shows poses in consecutive timesteps corresponding to the reference images on the left.}
\label{vibe_cmp}
\vspace{-3mm}
\end{figure*}

\section{Application}  \label{sec:application}
% \ruben{Maybe describe the tasks we are solving within the experiments section?}
% \jiaman{I think we should only show results in experiment section, and showsing the approach here. Like the paper learned optimize priors for 3d modeling.}

Our HM-VAE provides a generalized motion prior that can be applicable in various tasks like 3D video pose estimation, motion interpolation and motion completion. In this section, we introduce the applications we consider and describe the effective strategy used for each application. We provide qualitative and quantitative results in the next section.

\paragraph{3D Video Pose Estimation.}
% Our model consists of encoder and decoder.
% For refining video pose estimation results, we utilize both our learned encoder and decoder. Concretely, we take the pose estimation results as input to encoder, then we decode from the encoded latent vector. Our learned prior model is designed for a fixed window size, however, we aim to process arbitrary long motion sequences with our learned motion priors. Thus, we propose a sliding window strategy, enabling the motion restoration for arbitrary long sequence. Suppose our learned model is trained on window size with $K$ frames, by feeding $K$ poses denoted as $N_1, ..., N_K$ to encoder and then decode, we will get refined $K$ poses denoted as $M_1, ..., M_K$. To eliminate the discontinuity issue of two consecutive windows, for example $W_1 = {t_1, ..., t_K}$, $W_2 = {t_{K+1}, ..., t_{2K}}$, discontinuity happen between $t_K$ and $t_{K+1}$, we introduce a center frame strategy. We only take $\frac{K}{2}$th frame $M_{\frac{K}{2}}$ as refined final result, added to our final refined sequence $S$. And the window is shifted by one timestep along the input motion sequence for processing next window. For each window size of pose sequences, we formally defined the operations as follows
Our learned motion prior provides an effective strategy to refine video based pose estimations. Concretely, we take potentially noisy pose estimates as input to the encoder, then decode refined poses using the encoded latent vector.
Our HM-VAE is designed for a fixed window size of $T$ frames. In order to have our method process sequences of any length, we could simply partition the input sequence into windows of $T$ frames. However, with no overlap across the time windows, we observe that this may result in discontinuities. 
Therefore, we propose a sliding window strategy using center frames to process arbitrarily long sequences. Specifically, for each time window we process, we only update the pose of the center frame with the refined result and shift the time window one step. %\duygu{I don't think we need the following equations, what we do is clear?}  
%\jiaman{When we process the next window, we always take original input poses.}
% To eliminate the discontinuity, we introduce a center frame strategy.
We take the $\frac{T}{2}$th frame $M_{\frac{T}{2}}$ as the refined final result, added to our final refined sequence $S$.
And the window is shifted by one timestep along the input motion sequence for processing the next window.
For each window size of pose sequences, we formally define the process as follows, where $W_2$ represents next window. 

% $$ W_2 = {t_2, ..., t_{K+1}}$$.
% $$ z = Enc(N_1, ..., N_K)$$
% $$ M_1, ..., M_{\frac{K}{2}}, ..., M_K = Dec(z)$$
% $$ S = S + M_{\frac{K}{2}}$$

\begin{align}
z &= Enc(N_1, ..., N_T), \\
M_1, ..., M_{\frac{T}{2}}, ..., M_T &= Dec(z), \\
S &= S \cup M_{\frac{T}{2}} \\
W_2 &= {N_2, ..., N_{T+1}}
\end{align}

% \paragraph{Motion Completion}

\paragraph{Motion Interpolation and Completion.}
% For motion interpolation, given key frame poses, we aim to generate motion in-between by optimizing our hierarchical prior. For motion completion task, given only upper body joint rotations, we aim to generate complete motions. We discard the encoder part, and utilize the trained decoder to accomplish these tasks. The optimization objective is to minimize the reconstruction loss between given target and decoded poses. Similar to reconstruction loss used in our prior training, our reconstruction loss for optimization also consists of three terms including 6D rotation representation, rotation matrix and joint positions after forward kinematics layer. 
A common setup in motion synthesis is to generate motion sequences given a sparse set of keyframes, which we refer to as motion interpolation. Motion completion, on the other hand, focuses on synthesizing complete body motion from partial observations, e.g., completing the motion of the lower body by observing the upper body. For both motion interpolation and completion tasks, we simply utilize the decoder of HM-VAE to synthesize motion while searching for an optimal latent code to match the given observations (i.e., sparse keyframes or partial body motion). The optimization objective is to minimize the reconstruction error between the given observations and the corresponding decoded poses. We define the reconstruction objective as a combination of three terms including matching the joint rotations using both 6D rotation and rotation matrix representations and matching the joint positions after forward kinematics:
% $$ L_{rec} = L_{6d} + L_{rot} + \lambda_1 L_{pose}$$
\begin{equation}
    L_{rec} = L_{6d} + L_{rot} + \lambda_1 L_{joints}
\end{equation}
% We introduce mask $M$ to constrain the optimization with incomplete motions as objective,    
% \ruben{Can you elaborate why you introduce the mask?} \jiaman{mask for optimizing with given target},   
% $$ L_{6d} = \sum_{i=1}^{T} ||M(X'_t-X_t)||2$$
Concretely, we perform optimization in two phases. Starting with randomly sampled latent vectors $z_l$ and $z_g$, in the first phase, we optimize for the latent vectors that minimize $L_{rec}$ as the only objective. The decoder parameters $\theta$ are fixed during this phase. In the next phase, we optimize for the decoder parameters $\theta$~\cite{yang2020deep} while keeping the latent vectors fixed. In this second phase, we introduce a regularization loss to constrain $\theta '$ and prevent it from deviating too much from the pre-trained parameters $\theta$. Thus, the optimization objective in the second phase becomes:  
% $$ L_{opt} = L_{rec} + \lambda_2 ||\theta - \theta '||^2$$
\begin{equation}
    L_{opt} = L_{rec} + \lambda_2 ||\theta ' - \theta||^2
\end{equation}

\section{Experiments}
% Figures: 
% model
% Smooth curve
% Completion results
% Pose estimation results
In this section, we first describe the dataset we use for training and evaluation. Then we showcase the results of applying our HM-VAE in the applications we introduced in the previous section. Finally, we perform an ablation study to validate the effectiveness of our overall approach. 

\paragraph{Dataset.}
We use the AMASS dataset~\cite{mahmood2019amass} for training HM-VAE. AMASS dataset is a large collection of 15 motion capture datasets with a unified data representation. The dataset has more than 40 hours of motion data and serves as a great testbed for motion modeling.  
% \ruben{Elaborate more about the dataset. For people who don't know this dataset consists of many datasets, it will sound like you only have a single dataset}. 
We use the same validation and testing split introduced in VIBE~\cite{kocabas2020vibe}. 
For refining video based pose estimates, we use 3DPW~\cite{von2018recovering}, a 3D motion in the wild dataset, as our test set.  
For the motion interpolation task, we train our HM-VAE on the LAFAN1 dataset~\cite{harvey2020robust} to provide quantitative comparisons to the baseline methods. LAFAN1 consists of high-quality motion capture data with specific action types. We follow the data split proposed in \cite{harvey2020robust} and use subjects 1, 2, 3 and 4 as training, and subject 5 for testing.

\paragraph{Implementation Details.}
We use a batch size of $8$ for training. The KL divergence weight $\beta$ is set to $0.003$. Unless noted otherwise, we train HM-VAE with motion sequences of length 64. While training our HM-VAE, to prevent the learning dominated by either shallow or deep latent vector, we use similar strategy proposed in ~\cite{li2020progressive}. We first only train our model with deep latent vector, then start training both shallow and deep latent vectors after $50000$ iterations. 
% For the motion interpolation experiments, we perform $150$ optimization iterations, with $50$ iterations for the first phase and $100$ iterations for the second phase.
% For the motion completion experiments, we increase the optimization iterations to $300$ with $100$ iterations belonging to the first phase described in Section~\ref{sec:application}.
For the motion interpolation experiments, we found our method converged at around $150$ iterations of optimization, with $50$ iterations for the first phase and $100$ iterations for the second phase.
For the motion completion experiments, we found our optimization converged at around $300$ iterations with $100$ iterations belonging to the first phase.

% We empirically found that our method converged at around $150$ iterations for the motion interpolation task, and at around $300$ iterations for motion completion.
% \duygu{can we give an insight why we need more iterations for motion completion?}
% \jiaman{Can we say empirically? The iterations is decided by observing the reconstruction loss changes. It's a rough estimated value.}
% \ruben{Do you mean the reconstruction converges at 300 iterations and so we choose that number?}

% The batch size of motion prior training is 8. We set the weight of kl divergence loss as 0.003. For optimization in motion interpolation, we set the optimization iterations to 150 while 50 iterations belong to phase 1. For optimization in motion completion, we set the optimization iterations to 300 while 100 iterations belong to phase 1.      

\subsection{Results}
\paragraph{3D Video Pose Estimation.}

% We first show that our motion prior model can be applied on top of different 3D video pose estimation methods. By encoding the predicted results from other methods, we can get the latent vector and decode it back to smooth motions. We show quantitative results in Table~\ref{table:pose-estimate-res}.
% \duygu{Introduce the error metrics we use}
% \jiaman{Introduced in next paragraph}
In this section, we show that our model can be used to refine the results of off-the-shelf 3D video pose estimation methods. %By encoding the noisy 3D pose sequences, we project them into our latent space, and decode them back as plausible motions. 
In order to adapt HM-VAE to different global rotations and frame rate among different datasets, we train our HM-VAE with data augmentation.
% \ruben{Will this data augmentation help VIBE?}
% \jiaman{This will not help VIBE. VIBE is doing prediction from images. We use it because VIBE has different root rotation distribution from AMASS}.
% \ruben{Let's clarify this in the text. I am sure a reviewer may ask this during rebuttal so let's prevent it from happening}
Our data augmentation consists of different frame rates and random global rotations.
Also, we use the HM-VAE model trained with a window size 8 in this application which we observe has a better reconstruction quality.
% Note that we cannot apply our data augmentation to the 3D video pose estimation methods because they take images as input.

We show quantitative results in Table~\ref{table:pose-estimate-res} where we test our method with inputs obtained by both VIBE~\cite{kocabas2020vibe} and HumanDynamics (HD)~\cite{kanazawa2019learning}. We report errors using the same metrics as VIBE~\cite{kocabas2020vibe}. Specifically, we report the mean per joint position error with (PA-MPJPE) and without (MPJPE) the Procrustes-alignment, as well as the mean per joint acceleration and acceleration error. In Figure~\ref{vibe_cmp}, we show the acceleration error curves as well as example poses obtained for consecutive timesteps. Compared to current state-of-the-art methods, our refined motions have smaller acceleration errors. While previous approaches are prone to abrupt changes across consecutive poses as shown in Figure~\ref{vibe_cmp}, our model smooths out these noisy estimates. We refer the readers to the supplementary video for a detailed comparison.

\begin{table}[t!]
\small
\begin{center}
%\centering
\footnotesize{
%\resizebox{2\columnwidth}{!}{
\setlength{\tabcolsep}{5pt}
\begin{tabular}{@{}l||cccccc@{}} 
%\begin{tabular}{|c|c|c} 
 \hline
 & PA-MPJPE  & MPJPE & ACCEL& ACCER \\ \hline\hline
HD~\cite{kanazawa2019learning} & 72.17 & 115.97 & 14.96 & 14.73 \\ \hline
HD~\cite{kanazawa2019learning} w Prior & \textbf{71.39} & \textbf{113.90} & \textbf{5.21} & \textbf{8.36} \\ \hline
VIBE~\cite{kocabas2020vibe} & 56.56 & 93.59 & 27.12 & 27.99  \\ \hline
VIBE~\cite{kocabas2020vibe} w Prior & \textbf{55.84} & \textbf{92.43} & \textbf{6.03} & \textbf{9.15}  \\ \hline
\end{tabular}
}
\end{center}
\vspace{-4mm}
\caption{\small 3D Video Human Pose Estimation Results in 3DPW Testing Dataset. }
\label{table:pose-estimate-res}
\vspace{-4mm}
\end{table}

% The prediction from partial-body work may not be accurate which influence the quantitative evaluation of our method. Thus, we also showcase results with 3DPW partial-body ground truth joints as target.  

% \begin{table}[t!]
% \small
% \begin{center}
% %\centering
% \footnotesize{
% %\resizebox{0.8\columnwidth}{!}{
% \setlength{\tabcolsep}{10pt}
% \begin{tabular}{@{}l||ccccccc@{}} 
% %\begin{tabular}{|c|c|c} 
%  \hline
%  & PA-MPJPE & MPJPE & ACCEL& ACCER    \\ \hline\hline
% Partial-body & 90.24 & 126.97 & 43.30 & 45.21  \\ \hline
% Our Recovery  & \textbf{89.37} & \textbf{125.20} & \textbf{9.77} & \textbf{16.62}  \\ \hline
% % Given Partial GT  & - & - & -    \\ \hline
% \end{tabular}
% }
% \end{center}
% \vspace{-1mm}
% \caption{3D Video Pose Estimation in Cropped 3DPW Evaluation}
% \label{table:partial-pose-estimate-res}
% \vspace{-1mm}
% \end{table}
\begin{figure*}[h]
% \vspace{-2mm}
\begin{center}   \includegraphics[width=0.85\linewidth]{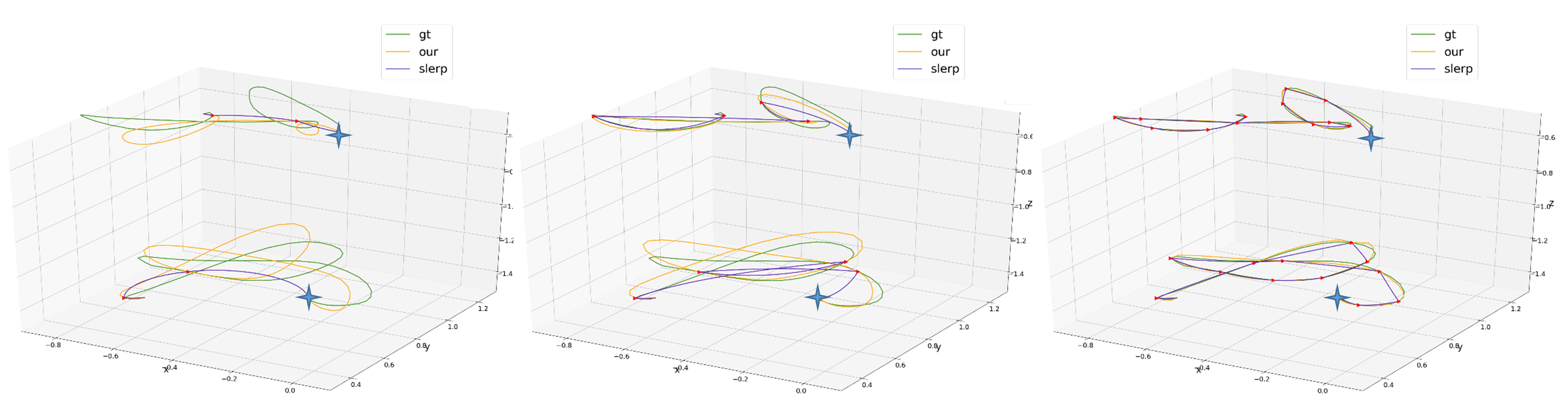}
\end{center}
\vspace{-4mm}
\caption{Local trajectory comparison for motion interpolation in AMASS data. From left to right is the trajectory when key frame interval is 30, 15, 5 respectively. The upper curves represent the left wrist, the lower curves represent the right ankle. The star symbol represents the starting point, the arrow symbol represent the position of key frames. Our results show similar moving patterns to ground truth, while Slerp differs a lot when key frame interval is large. }
\label{motion_interp_local_trajectory}
% \vspace{-3mm}
\end{figure*}

\begin{figure*}[h]
\begin{center}
% \vspace{-5mm}
% \includegraphics[width=100pt,height=25pt]{figures/model_pipeline.pdf}
\includegraphics[width=0.8\textwidth]{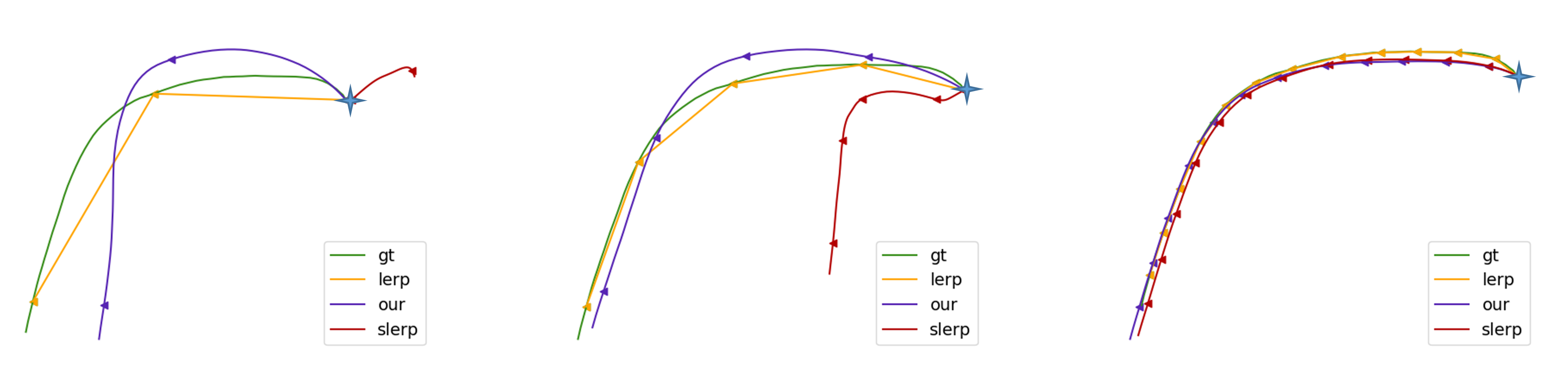}
\end{center}
\vspace{-4mm}
\caption{Global root trajectory comparison for motion interpolation in AMASS data. From left to right is the trajectory (in xy plane) when key frame interval is 30, 15, 5 respectively. The star symbol represents the starting point, the arrow symbol represents the position of key frames. 
}
\label{motion_interp_global_trajectory}
\vspace{-1mm}
\end{figure*}

\begin{figure}[t!]
\vspace{-2mm}
\begin{center}   \includegraphics[width=\linewidth]{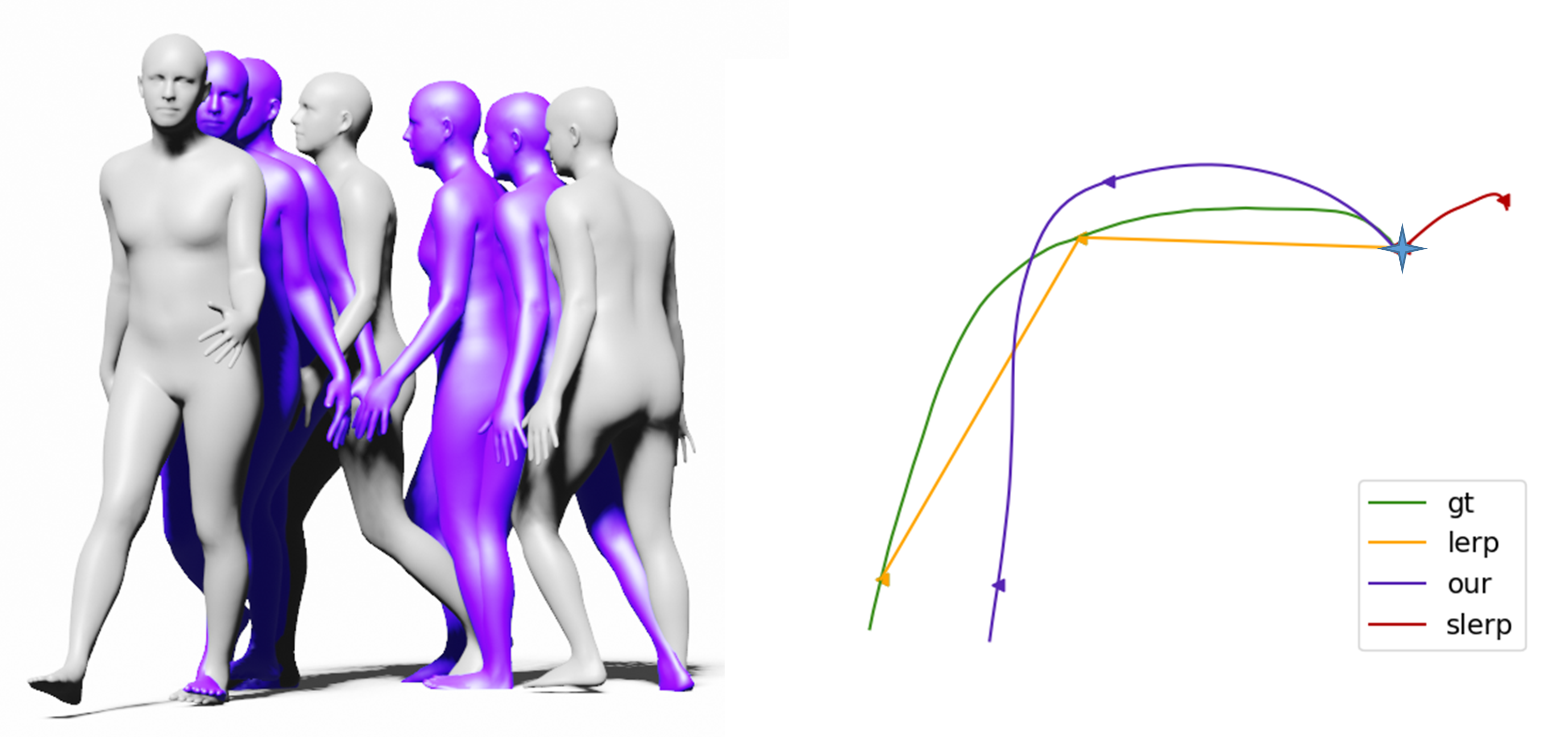}
\end{center}
\vspace{-5mm}
\caption{Motion interpolation results in AMASS test data. The gray mesh shows key frame poses, the purple mesh show the generated poses. The interval between two key frames is 30 frames. The right figure shows the global trajectory comparison for this motion sequence.}
\label{motion_interp_mesh_vis}
\vspace{-4mm}
\end{figure}

\vspace{-4mm}
\paragraph{Motion Interpolation.}
% \ruben{We need more description about how the baseline results are computed}
In order to demonstrate the effectiveness of HM-VAE for the motion interpolation task, we compare it to appropriate baseline methods. Specifically, in order to interpolate local joint rotations, we use the standard spherical linear interpolation (Slerp). Since interpolation quality is directly related to the number of missing frames, we perform our evaluations in four settings where $5, 15, 30, 45$ frames are missing in each setting. Following the same setting as ~\cite{harvey2020robust}, given the starting $10$ frames and ending $1$ frame as key frames, we aim to generate the frames in-between. We show quantitative comparisons in terms of local pose estimation in Table~\ref{table:lafan_interpolation}. In addition to the metrics introduced before, we also report the global quaternion loss proposed by the original benchmark~\cite{harvey2020robust}. We show that our method outperforms the Slerp baseline quantitatively.
We also show that the performance achieved by our human motion prior is competitive against the in-betweening specific method from~\cite{harvey2020robust} in the global quaternion loss metric.
Please note that we use the LAFAN1 dataset for this evaluation to compare against the global quaternion errors directly reported by~\cite{harvey2020robust} since their code is not published and the authors were not able to run their model on our dataset.  
% \jiaman{Note that figure 4, 5, 6 are all in AMASS data}
We also provide additional qualitative results in the AMASS dataset. We visualize local joint trajectories in Figure~\ref{motion_interp_local_trajectory} for a walking motion sequence.
Our results preserve the original motion patterns while Slerp fails to model the local motion when the interval between two key frames becomes large.
We further demonstrate global trajectory interpolation with our method and the alternatives.
Specifically, we use our global trajectory estimation module by providing the local motion predicted by our method as well as Slerp. In addition, we also define a simple baseline where we linearly interpolate the global root position of the sparse keyframes (lerp).  As shown in Figure~\ref{motion_interp_global_trajectory}, the trajectory estimated by our method more closely resembles the ground truth.      
We also show a mesh visualization result for motion interpolation in Figure~\ref{motion_interp_mesh_vis}. For more qualitative results, we encourage readers to check our accompanying video.

\begin{figure*}[h]
\begin{center}
\vspace{-5mm}
\includegraphics[width=0.8\textwidth]{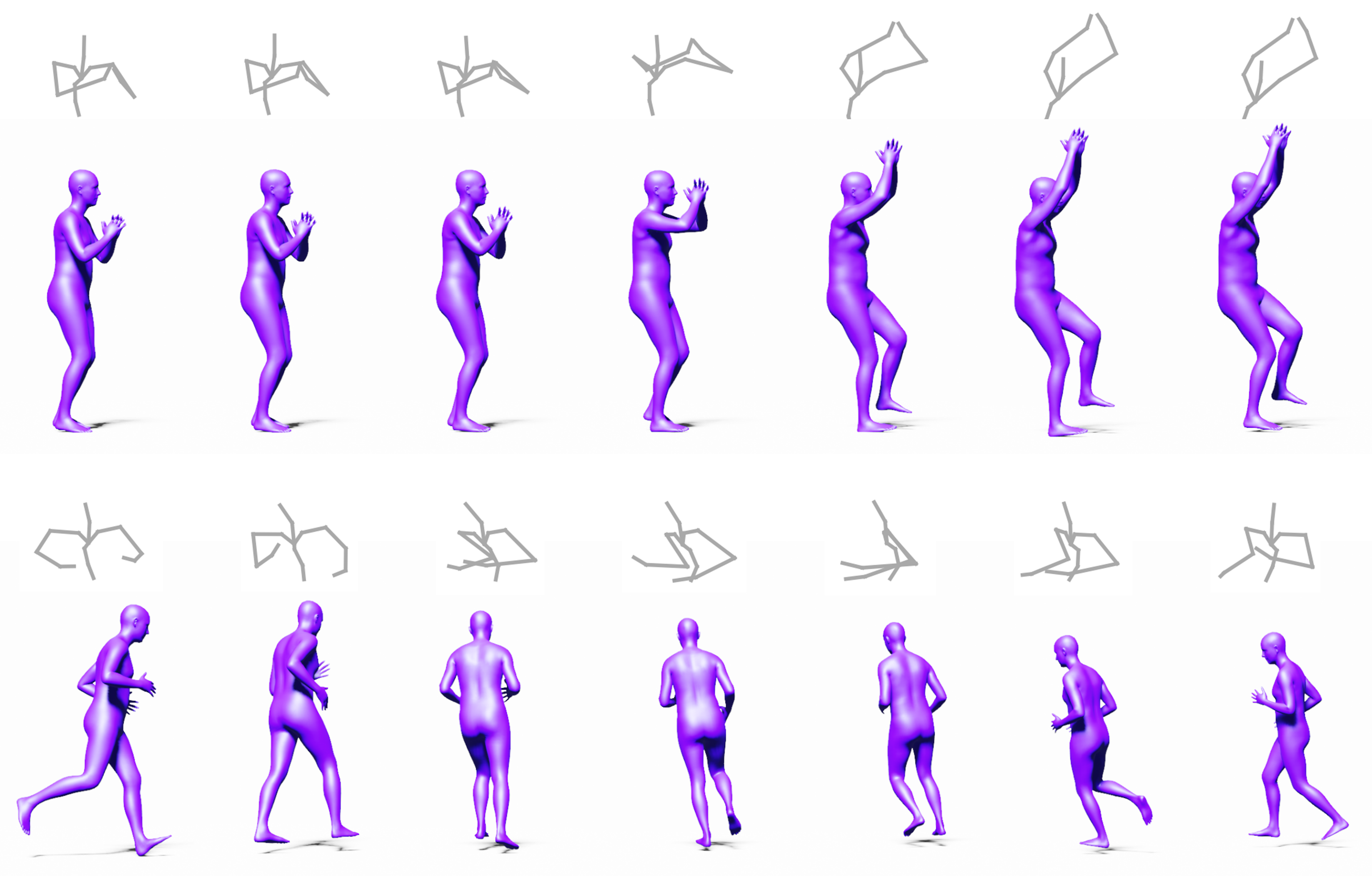}
\end{center}
\vspace{-6mm}
\caption{Motion Completion Results. Given upper body joint rotation as optimization objective, the prior model can complete whole motion sequences.}
\label{motion_completion}
\vspace{-6mm}
\end{figure*}

\begin{table}[t!]
\small
\begin{center}
%\centering
\footnotesize{
%\resizebox{2\columnwidth}{!}{
\setlength{\tabcolsep}{10pt}
\begin{tabular}{@{}l||cccccc@{}} 
%\begin{tabular}{|c|c|c} 
 \hline
 & 5  & 15 & 30 & 45 \\ \hline\hline
MPJPE-Slerp & 16.02 & 57.13 & 96.54 & 118.96 \\ \hline
MPJPE-Ours & \textbf{14.08} & \textbf{45.09} & \textbf{90.41} & \textbf{117.93} \\ \hline \hline
PAMPJPE-Slerp & 15.82 & 54.11 & 83.66 & 92.4 \\ \hline
PAMPJPE-Ours & \textbf{12.03} & \textbf{38.37} & \textbf{72.21} & \textbf{86.06} \\ \hline \hline
ACCEL-Slerp & 1.75 & 0.78 & 0.35 & 0.23 \\ \hline
ACCEL-Ours & 4.79 & 4.48 & 4.08 & 3.61 \\ \hline \hline
ACCER-Slerp & 5.98 & 5.97 & \textbf{6.05} & \textbf{6.06} \\ \hline
ACCER-Ours & \textbf{5.31} & \textbf{5.83} & 6.54 & 6.75 \\ \hline \hline
Global Quat-Slerp & 0.22 & 0.62 & 0.98 & 1.25 \\ \hline
% Global TG rec & 0.21 & 0.48 & 0.83 & 1.20 \\ \hline
% Global TG complete & 0.17  & 0.42 & 0.69 & 0.94 \\ \hline
Global Quat-\cite{harvey2020robust} & \textbf{0.17}  & \textbf{0.42} & \textbf{0.69} & \textbf{0.94} \\ \hline
Global Quat-Ours & 0.24 & 0.54 & 0.94 & 1.25 \\ \hline
\end{tabular}
}
\end{center}
\vspace{-5mm}
\caption{\small Quantitative Evaluation for Motion Interpolation in LAFAN1 Dataset.}
\label{table:lafan_interpolation}
\vspace{-6mm}
\end{table}

\vspace{-4mm}
\paragraph{Motion Completion.}
% \ruben{Can we have more discussion about the results here?}
Given only upper body joint rotations as target, we aim to recover the complete body motion sequences.
For this experiment, we use motion sequences from the testing and validation split of the AMASS dataset.
As shown in Figure~\ref{motion_completion}, our approach is able to restore complete motions since the global latent space capture the correlations among different joints.
Therefore, the missing lower legs movement that matches the given upper body is retrieved from the learned latent space for human motion.

\subsection{Ablation Study}
\vspace{-2mm}
% \ruben{More discussion?}
% \jiaman{what kind of discussion? I have some numbers in supplemental.tex, not sure if we need to put them here}
In order to motivate the design choices we made, we perform an ablation study where we compare our HM-VAE with a non-hierarchical motion VAE (M-VAE) and a VAE with only temporal convolution layers (TCN-VAE). The temporal convolution layers were used in training an autoencoder for motion processing~\cite{holden2015learning,holden2016deep}.
We compare our model to the alternatives in the task of motion reconstruction using the AMASS dataset. 
Specifically, for each testing sequence, we take the local joint rotations as input to the encoder and then decode the motion from the mean vector.
We measure the mean joint reconstruction error as shown in Table~\ref{table:motion_rec}.
Our HM-VAE model outperforms the M-VAE by a large margin in motion reconstruction evaluation.
And the model with skeleton-aware architecture has superior performance than its temporal convolution counterpart.
Therefore, we show that skeleton operations from the skeleton-aware architecture are important for modeling the human body structure in comparison to using standard temporal convolution.
Moreover, modeling a global and local motion latent spaces further improve the human motion modeling power of the skeleton-aware architecture. 
% \ruben{Do we have numbers for pose estimation using the non-hierarchical prior?} \jiaman{We have that number in supplemental.tex, but I think we don;t need that number in the paper? since we are showing we do better in ablation study, then we just use the same model for all the applications. }
\\

\begin{table}[t!]
\small
\begin{center}
%\centering
\footnotesize{
%\resizebox{2\columnwidth}{!}{
\setlength{\tabcolsep}{8pt}
\begin{tabular}{@{}l||cccccc@{}} 
%\begin{tabular}{|c|c|c} 
 \hline
 & PA-MPJPE  & MPJPE & ACCEL & ACCER \\ \hline\hline
% 2 layer & 59.54 & 46.76 & 2.33 & 5.94 \\ \hline
TCN-VAE & 87.27 & 103.60 & 1.66 & 6.46 \\ \hline 
M-VAE & 59.71 & 74.34 & 2.36 & 6.15 \\ \hline
% HM-VAE & \textbf{46.94} & \textbf{60.85} & 2.45 & \textbf{6.08} \\ \hline
HM-VAE & \textbf{45.82} & \textbf{58.46} & 2.29 & \textbf{5.98} \\ \hline
\end{tabular}
}
\end{center}
\vspace{-5mm}
\caption{\small Motion Reconstruction Results in AMASS test data.}
\label{table:motion_rec}
\vspace{-6mm}
\end{table}

\vspace{-4mm}
\section{Conclusion}
\vspace{-2mm}
%\ruben{We should discuss the limitations of our method. We should also talk about potential future work based on the discussed limitations. A potential future work I can see is fine-tuning our model for specific human motion tasks. Something like few-shot motion [[insert task]] type of extension.}
%\jiaman{Limitation: sampled motion sequences are not realistic enough; the restored motion results cannot guarantee foot stick to floors; accumulation errors of trajectory prediction model;
%Future work: environment-aware, consider foot contact, environments like stairs; add differentiable physics simulator for more realistic motions? condition on action, with intentions;}
We propose a task-generic motion prior using a hierarchical motion VAE. We demonstrate the effectiveness of the prior in various applications including 3D video pose estimation, motion interpolation, and motion completion. By learning a global and local embedding, our prior can faithfully model human motion. While our prior enables to refine video-based human motion estimation results by reducing jitters, it also performs on-par with task specific methods for motion interpolation and completion. There are some limitations of our method we would like to address in future work. 
We observe that there are accumulation of errors when predicting the global trajectory for a long sequence. Exploring more constraints like foot contact during both training and inference might be a potential approach to address this. 
% Exploring scheduled sampling techniques during training might be a potential approach to address this. 
While we show that our prior is effective in different applications, using few-shot learning to better adapt to specific tasks is another interesting direction. Finally, incorporating certain physical properties and action conditions are also promising directions.

\paragraph{Acknowledgements.}
This research was conducted at Adobe, at University of Southern California and USC Institute for Creative Technologies. Research was sponsored by the Army Research Office and was supported under Cooperative Agreement Number W911NF-20-2-0053, and sponsored by the U.S. Army Research Laboratory (ARL) under contract number W911NF-14-D-0005, the CONIX Research Center, one of six centers in JUMP, a Semiconductor Research Corporation (SRC) program sponsored by DARPA; and in part by the ONR YIP grant N00014-17-S-FO14. Affiliation with Pinscreen and the University of California at Berkeley was supported by DARPA under cooperative agreement HR00112020054. The views and conclusions contained in this document are those of the authors and should not be interpreted as representing the official policies, either expressed or implied, of the Army Research Office or the U.S. Government. The U.S. Government is authorized to reproduce and distribute reprints for Government purposes notwithstanding any copyright notation.

{\small
\bibliographystyle{ieee_fullname}
\bibliography{egbib}
}

\end{document}